\documentclass[journal]{IEEEtran}

\usepackage{xcolor,soul,framed} %,caption

\colorlet{shadecolor}{yellow}
\usepackage[pdftex]{graphicx}
\graphicspath{{../pdf/}{../jpeg/}}
\DeclareGraphicsExtensions{.pdf,.jpeg,.png}

\usepackage[cmex10]{amsmath}
\usepackage{tabularx}
\usepackage{adjustbox}
\usepackage{amsmath,amssymb,amsfonts}
\usepackage{textcomp}
\usepackage{hyperref}
\usepackage{url}
\usepackage{adjustbox}
\usepackage{physics}
\usepackage{multirow}
\usepackage{xcolor}
\usepackage[utf8]{inputenc}
\usepackage[english]{babel}
\usepackage{soul}
\usepackage{algorithm}
\usepackage{float}
\usepackage{arevmath}     % For math symbols
\usepackage{algpseudocode}

\begin{document}
    \title{OpTorch: Optimized deep learning architectures for resource limited environments}
  \author{Salman Ahmed, \and Hammad Naveed, \and \IEEEmembership{CBRL, Department of Computer Science, \\ NUCES-FAST, Islamabad}% <-this % stops a space

  \thanks{This work was supported by the National Center in Big Data and Cloud Computing (NCBC) and the National University of Computer and Emerging Sciences (NUCES-FAST), Islamabad, Pakistan.}}

% ====================================================================
\maketitle

% === ABSTRACT ====================================================================
% =================================================================================
\begin{abstract}
Deep learning algorithms have made many breakthroughs and have various applications in real life. Computational resources become a bottleneck as the data and complexity of the deep learning pipeline increases. In this paper, we propose optimized deep learning pipelines in multiple aspects of training including time and memory. 
OpTorch is a machine learning library designed to overcome weaknesses in existing implementations of neural network training. OpTorch provides features to train complex neural networks with limited computational resources.
OpTorch achieved the same accuracy as existing libraries on Cifar-10 and Cifar-100 datasets while reducing memory usage to approximately 50\%. We also explore the effect of weights on total memory usage in deep learning pipelines. In our experiments, parallel encoding-decoding along with sequential checkpoints results in much improved memory and time usage while keeping the accuracy similar to existing pipelines. OpTorch python package is available at available at \url{https://github.com/cbrl-nuces/optorch}.
\end{abstract}

% === KEYWORDS ====================================================================
% =================================================================================

% === I. INTRODUCTION =============================================================
% =================================================================================
\section{Introduction}

\IEEEPARstart{N}{eural Networks} (NN) are mathematical models designed to mimic information processing of the human brain. Generally, NN consists of different layers of neurons which communicate with each other to make a decision. Neurons and number of layers are key components to design different architectures. NN learns to activate neurons based on inputs passed to it \cite{b13}.

Deep Neural Networks have shown improved performances on a wide variety of applications \cite{b14}. Typically, growing the size and complexity of neural networks results in improved accuracy. As the model size grows, memory and time to train such networks also increase. One of the greatest achievements of deep learning in 2020 was by OpenAI to introduce GPT-3. The main takeaway from the study is that complex and large models can solve complex problems in language modeling\cite{b2}. 

This paper proposes two different optimizations: a) Gradient-flow and b) Data-flow. Gradient-flow optimization represents optimization with in the architecture while Data-flow optimization represents optimization out of the NN architecture including data passage, selection and augmentation. 

There are multiple existing Gradient-flow optimizations like Mixed-precision training \cite{b1}. Mixed-precision implementation uses only half-precision format to save limited float numbers in comparison to single-precision format. Mixed precision uses both 16-bit and 32-bit floating-point types during the model training. This makes it run faster and uses less memory. This technique keeps certain parts of the model in the 32-bit types for numeric stability. 

Sequential models are executed in a sequence of lists of layers called segments. Therefore, we can execute sequential models in segments and checkpoint each segment. All segments except the last will execute in a way that they will not store the intermediate activations. The inputs of each segment will be saved for re-running the segment in the backward pass. 

Existing implementations store output of intermediate activations before calculating gradients. In some situations, for example, if NN is small enough to fit in memory, it is better to store the output of the activations before calculating the gradients in order to speed up learning. 
Other existing techniques that optimize neural networks include small Mini-batch training alongside Batch Accumulation. Some layers or functions in neural networks like batch-normalization works better with large batch size. Lin {\it et al.} trained NNs using small mini-batches and accumulation of gradients to show effect of batch accumulation \cite{b9}. 

To solve complex problems we need deep neural networks large enough to understand the problem. Existing standard implementations cannot be used to train large networks with limited computational resources. 

OpTorch provides Mixed-precision flexibility as well as Sequential-checkpoints.  Instead of saving all activation outputs, OpTorch stores limited number of activations and recomputes remaining at run-time. 
OpTorch provides multiple Data-flow optimizations as well. 
We propose an idea to generate batches in parallel similar to Chen {\it et al.} \cite{b3} however, we encode and compress batches before passing to NN to save memory and passage time up-to 16X. Such compression reduces at-least 20\% training time.

We performed training experiments with enhanced versions of famous architectures including Resnets (18, 34, 50) \cite{b4}, EfficientNets (B0, B1, B2, B3, B4, B5, B6, B7) \cite{b5}, and Inception-V3 \cite{b6} on Cifar-10 and Cifar-100 datsets using OpTorch.

% === Data-flow Optimization ========================
% =================================================================================

\section{Methods}
\subsection{Image Data-flow Optimization}
We designed an improved representation of image data. We encode multiple images to one matrix of the same size and pass this matrix to the network by saving up-to 16X memory.
OpTorch contains a custom layer for NN to decode such encoded images. This enhancement in the pipeline allows us to have unique data selection methods like selective-batch-sampling (SBS) to augment data and improve the training time.

\subsubsection{Encoding and Selective-batch-sampling(SBS)}

On-the-fly pre-processing for specific classes can be performed using SBS, meaning specific number of images for a specific class with in each batch can be selected. It can be considered as controlling each batch and ratio of each class in a batch while training.
Encoding images allows us to perform pre-sampling and controlled augmentation. The pixels in an image range between 0 - 255 values. Same positional pixel of N images can be encoded by using the equation below.
\[\sum_{i = 1}^N 256^i * M[i]\]
where M represents matrix of image, N represents number of images, and i represents location of pixel in image M.

Batch controlling will allow us to apply specific augmentations on specific classes. It will allow us to apply state of the art augmentations like MixUp \cite{b11}, CutMix \cite{b12} and AugMix \cite{b10} easily on specific combination of classes. Algorithm 1 explains the flow to encode a batch of images into a single matrix of the same size.

\begin{algorithm}
\caption{Encode}
\label{algo:encode_algo}
\begin{algorithmic}
\State Initialize $X$       \Comment{X : Batch of images (HxWxCxB)}
\State Initialize $A$       \Comment{A : Empty array of size HxWxC.}
\State Initialize $Z$       \Comment{Number of Images. ($Z \leq 16$)}
\State Initialize $i = 0$  
\While{$i \not= Z$}
\State $img = X[i]$
\State $domain = 256^{i}$
\State $A = A + (img * domain)$
\State $i = i + 1$
\EndWhile
\end{algorithmic}
\end{algorithm}

Encoding of images created new possibilities of SBS and image augmentations. Based on this encoding, we provide a pipeline for SBS which allows each batch to have a number of images with respect to class weights.

SBS allows pre-processing data differently for each class during training. Algorithm 2 explains the flow of SBS in our pipeline. 
It starts by initializing a variable with number of examples for each class in a batch. It selects specified number of examples for each class in each batch. This process allows us to pre-process each class differently in each batch.

\begin{algorithm}
\caption{Selective-batch-sampling}
\label{algo:sampling}
\begin{algorithmic}
\State Initialize $UC$       \Comment{UC : Unique Classes.}
\State Initialize $N$       \Comment{N : Number of Unique Classes.}
\State Initialize $W$       \Comment{W : Class Weights}
\State Initialize $X$       \Comment{X : Array of images (HxWxCxT)}

\State Initialize $Z$       \Comment{Number of Images. ($Z \leq 16$)}

\State Initialize $i = 0$  
\While{$i \not= N$}
\State Select subset of data for class $UC[i]$
\State select W[i] * Batch Size examples for batch
\State pre-process \& dump in batch
\State $i = i + 1$
\EndWhile
\end{algorithmic}
\end{algorithm}

\subsubsection{Decoding}

We designed a custom deep learning layer to decode each input matrix to original images. Algorithm 3 explains the flow to decode a batch of images into a single matrix of the same size. We calculate the modulus of each pixel by 256 then set the previous input matrix to divide by 256 to get a new image.

\begin{algorithm}
\caption{Decode}
\label{algo:decode_algo}
\begin{algorithmic}
\State Initialize $A$       \Comment{A : Encoded array of size HxWxC.}
\State Initialize $X$       \Comment{X : Empty Batch of images (HxWxCxB)}
\State Initialize $Z$       \Comment{Number of Images. ($Z \leq 16$)}
\State Initialize $i = 0$  
\While{$i \not= Z$}
\State $X[i] = A \mod 256$
\State $A = A\ \mathbf{div}\ 256$ \Comment{Division is Integer Division.}
\State $i = i + 1$
\EndWhile
\end{algorithmic}
\end{algorithm}

\subsubsection{OpTorch Loss-less Forced Encoding Option}
Loss-less encoding is an extended version of Algorithm 1. The domain of pixels in an image is between 0 - 255. If we divide each pixel by 2, the domain becomes 0–128, but will result in information loss. To convert this into loss-less encoding, we'll keep an offset based on the condition that the pixel was odd or even. This offset will help to decode the original pixel value. For float-64 datatype, Algorithm 1 can only encode 16 images, but Algorithm 4 can be used to encode 32 images in float-64 datatype and 32-bit for offset.

\begin{algorithm}
\caption{Loss-less Forced Encoding}
\label{algo:loss_less_encode_algo}
\begin{algorithmic}
\State Initialize $X$       \Comment{X : Batch of images (HxWxCxB)}
\State Initialize $A$       \Comment{A : Empty array of size HxWxC.}
\State Initialize $O$       \Comment{O : Boolean Empty array of size HxWxCxB.}
\State Initialize $Z$       \Comment{Number of Images. ($Z \leq 32$)}
\State Initialize $i = 0$  
\While{$i \not= Z$}
\State $img = X[i]$
\State $offset = img \mod 2$
\State $domain = 128^{i}$
\State $A = A + (img * domain)$
\State $O[i] = offset$
\State $i = i + 1$
\EndWhile
\end{algorithmic}
\end{algorithm}

\subsubsection{OpTorch Parallel Encoding-Decoding (E-D)}
\begin{figure*}
 \centering
\includegraphics[width=\textwidth,]{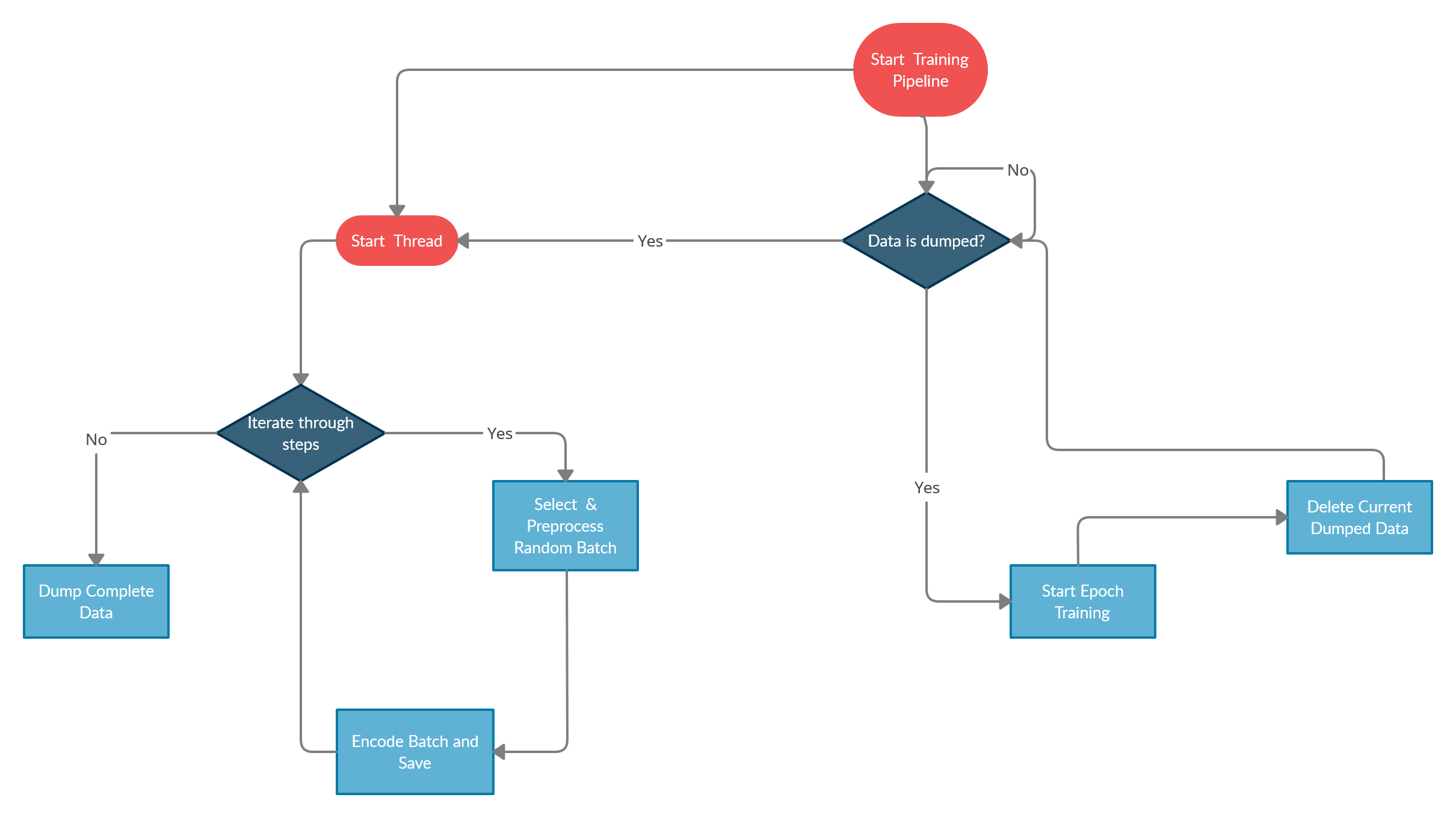}
  \caption{Figure represents an optimized flow of training pipeline. Flow starts with a if/else statement that data is dumped in encoded form or not. If not dumped then a thread will start to encode batches and perform all the pre-processing/augmentation and dump. Training will start after data is dumped for first time. In parallel of training a new thread will start to encode batches for next epoch and dump these encoded batches.}
  \label{fig:optim_pipeline}
\end{figure*}

\begin{figure*}
 \centering
\includegraphics[width=\textwidth,]{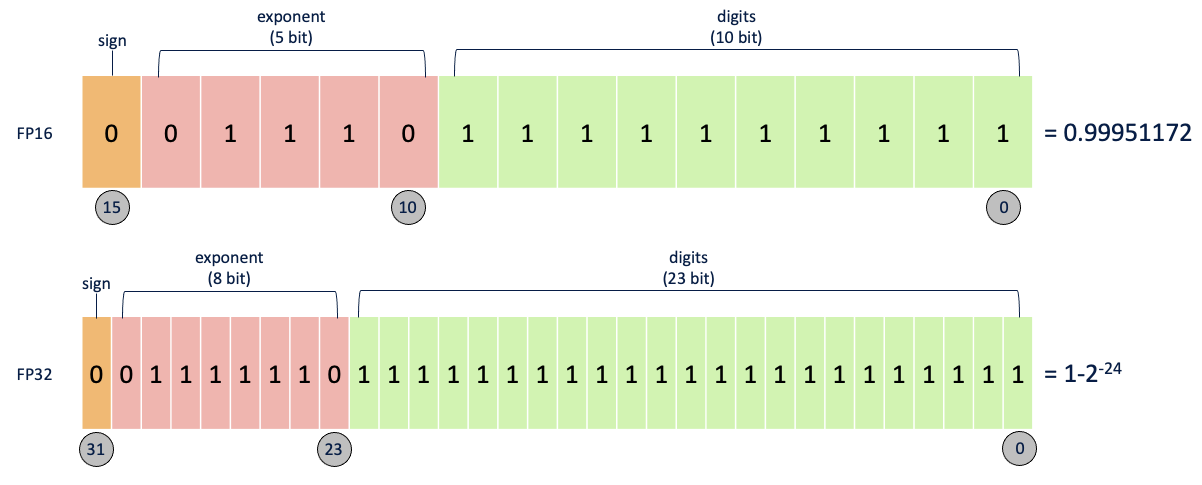}
  \caption{Figure represents comparison of FP16 (half precision floating points) and FP32 (single precision floating points). }
  \label{fig:fp}
\end{figure*}

We adjusted the training pipeline by introducing a new thread to encode batches in parallel. While training an epoch a thread will shuffle images, encode batches and prepare input for the next epoch. Figure \ref{fig:optim_pipeline} explains pipeline of training and encoding batches in parallel to improve training time.

\subsection{Gradient-flow Optimization}
Gradient-flow optimization represents optimization within an architecture. OpTorch provides multiple gradient-flow optimization features like Mixed-precision training, Sequential-checkpoints and recommendations for sequential-checkpoints.

\subsubsection{Mixed-precision training}
The technical standard used to represent floating point numbers in binary formats is IEEE 754, established in 1985 by the Institute of Electrical and Electronics Engineering.
\begin{figure}
 \centering
\includegraphics[width=0.5\textwidth,]{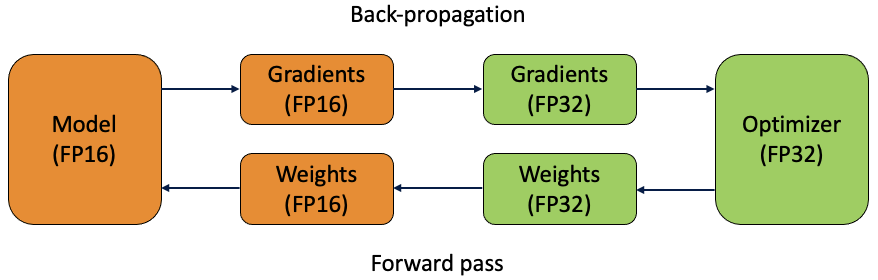}
  \caption{Figure represents mechanism of Mixed-precision training. FP16 weights are converted to FP32 before calculating loss and gradients and then converted back to FP16 to updated weights.}
  \label{fig:fp_conv}
\end{figure}
Traditionally, single precision (FP32) is used to represent parameters in deep learning. In FP32 format, 1 bit is reserved as the sign bit, 8 bits for the exponent (-126 to +127) and 23 bits for numbers. In Half precision (FP16), 1 bit is reserved as the sign bit, 5 bits for the exponent (-14 to +14), and 10 bits for the numbers  as shown in Figure \ref{fig:fp}.

In standard training, FP32 is used to represent model parameters, increasing memory usage significantly. In Mixed-precision training, FP16 is used to save model weights but precision decreases which results in low accuracy. Mixed-precision handles this effect on accuracy by converting FP16 weights to FP32 before loss and gradients calculations but stores weights in FP16 format as shown in Figure \ref{fig:fp_conv}.

\subsubsection{Sequential-checkpoint training}

\begin{figure}
  \includegraphics[width=0.5\textwidth,]{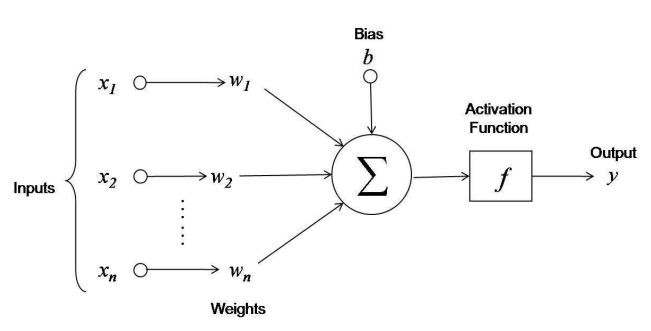}
  \caption{Operations done by Neuron.}
  \label{fig:neuron}
\end{figure}

A sequential neural network can be represented as stack of layers of neurons. Each layer can be represented as cluster of neurons. 
A neuron takes input and perform some mathematical operations to produce an output.
Figure \ref{fig:neuron} represents mathematical function of neuron, considering we have 3 inputs [x1, x2, x3] which is multiplied by weights [w1, w2, w3] and added to biases [b1, b2, b3]. After these operations neuron applies a non-linear transformation called activation function to obtain a value. 

This activation function is usually used to convert values to a range of 0 to 1 to help neural network learn.

\begin{figure}
  \includegraphics[width=0.5\textwidth,]{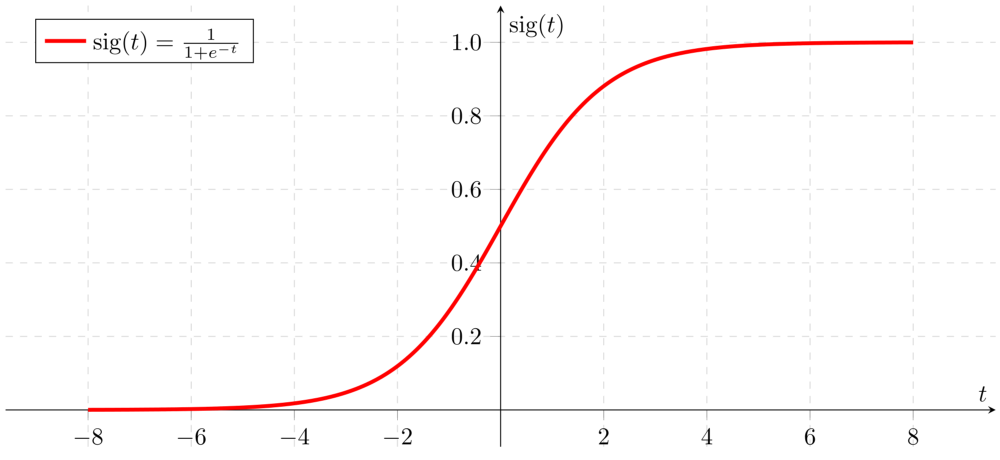}
  \caption{Sigmoid activation function.}
  \label{fig:activation}
\end{figure}

Figure \ref{fig:activation} represents a graph of a non-linear activation function called sigmoid. There are other activation functions available as well \cite{b15}.

\begin{figure}
  \includegraphics[width=0.5\textwidth,]{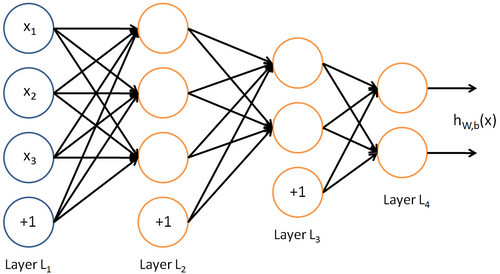}
  \caption{Neural Network Representation.}
  \label{fig:ann}
\end{figure}

Collection of such neurons on the same level is called a layer. Stack of such layers form a neural network. Figure \ref{fig:ann} represents a simple neural network. It can be read from left to right. Left most layer is called input layer. Middle 2 layers are called hidden layers. Last layer produces all possible outputs. +1 represent biases in layers. 

Connections between neurons have some weights. Back-propagation is used to train these weights by calculating partial derivative w.r.t loss/error. The partial derivative w.r.t loss/error provides the change in weight W to minimize loss/error. 

In practice, a NN of size of few hundred MBs can crash a GPU in the training process. Extra memory usage comes from following 2 main reasons to store extra information while training.
\begin{itemize}
  \item Necessary information to back-propagate (gradients of intermediate activations with respect to loss).
  \item Necessary information to calculate gradients.
\end{itemize}

\begin{figure}
  \includegraphics[width=0.5\textwidth,]{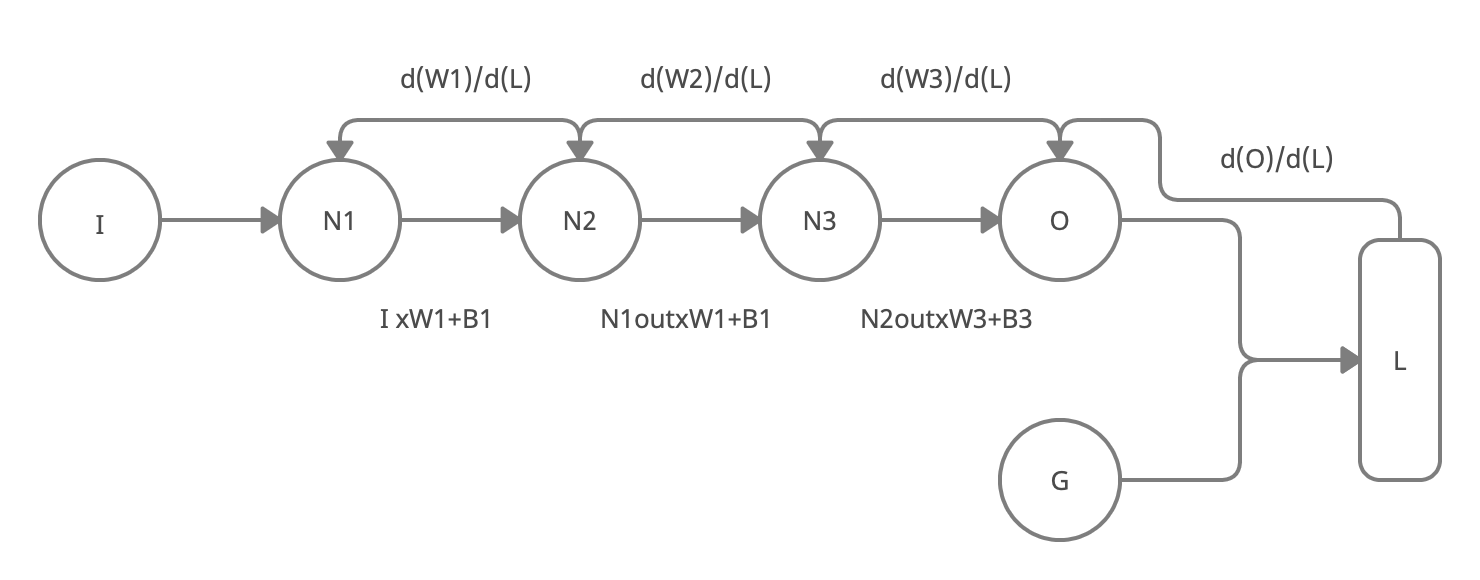}
  \caption{Flow of feed-forward and back-propagation in neural network.} 
  \label{fig:backprop}
\end{figure}

Both steps are essential as outputs of intermediate activations are required to apply chain rule but implementations can be flexible. To elaborate back-propagation and memory leakage consider a simple NN as shown in Figure \ref{fig:backprop}. 

Flow of back-propagation starts by calculating derivative of O w.r.t L. After this we calculate derivative of W3 w.r.t L. Derivative of W3 w.r.t L cannot be calculated directly because L is not directly dependent on W3. We calculate derivative of W3 w.r.t L in form of following equation using chain-rule.

\[
\dv{W3}{L} = \dv{W3}{N3out} * \dv{N3out}{O} * \dv{O}{L}
\]

Similarly, derivative of W2 w.r.t L is calculated in the form of following equation.

\[
\dv{W2}{L} = \dv{W2}{N2out} * \dv{N2out}{W3} * \dv{W3}{N3out} * \dv{N3out}{O} * \dv{O}{L}
\]

Weights of neural network are required but all other necessary information required to calculated gradients in back-propagation like N2out, N3out etc can be calculated at run-time.
All existing implementations in standard libraries save all these variables like N1out, N2out, and N3out and increase memory usage exponentially while doing forward pass and release memory after performing back-propagation. 

OpTorch provides a simple way to minimize memory usage. OpTorch creates checkpoint in intermediate outputs. Instead of saving all variables like N1out, N2out, and N3out we create a checkpoint and save N2out. For N3out calculation during back-propagation, a forward pass is done again from the previous checkpoint N2out and N3out is calculated directly. Similarly, we calculate N1out directly by doing partial forward pass from the first neuron. 
This can be scaled to a neural network with hundreds of layers. Instead of saving all outputs of these layers, we save a small number of checkpoints to control memory consumption. 

\section{Results}

Large deep neural networks are required to understand complex problems. Existing standard implementations cannot be used to train large DNN in resource limited settings. We propose OpTorch: A library that allows to train large neural networks in limited computational resources. We compare features of OpTorch with existing implementations in multiple environments based on memory consumption and time.

\begin{figure}
  \includegraphics[width=0.5\textwidth,]{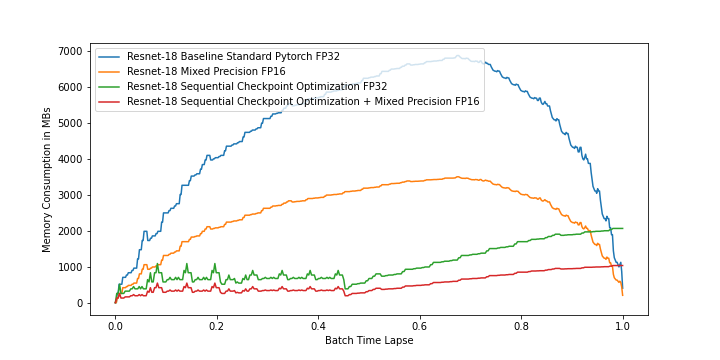}
  \caption{GPU Memory usage in 1 iteration. Batch for iteration consists of 16 images of size 512x512x3. } 
  \label{fig:mem_res}
\end{figure}

We first analyze GPU memory consumption of Resnet-18 architecture in 1 batch iteration. The batch consists of 16 images of size 512x512x3.
Figure \ref{fig:mem_res} represents analysis of memory consumption in training Resnet-18 with multiple enhanced pipelines. The x-axis corresponds to completion of 1 iteration from Resnet-18 whereas, the y-axis represents memory consumption in MBs. 

Creating checkpoints in Resnet-18 reduced memory consumption from 7000 MBs to just 2000 MBs. Such memory optimization comes with a cost. It takes more time to train by creating checkpoints than standard implementations because we have to do multiple sub-forward pass within a forward pass at run-time. 

To analyze performance of optimization methods on famous image models like Resnets (18, 50, 101), EfficientNets (B0, B1, B2, B3, B4, B5, B6, B7), and Inception-V3, we use CIFAR-10 dataset. CIFAR-10 comes with 60,000 color images of size 32x32x3 with 6,000 images per class, and 10 object classes.

\begin{figure*}
  \includegraphics[width=\textwidth,]{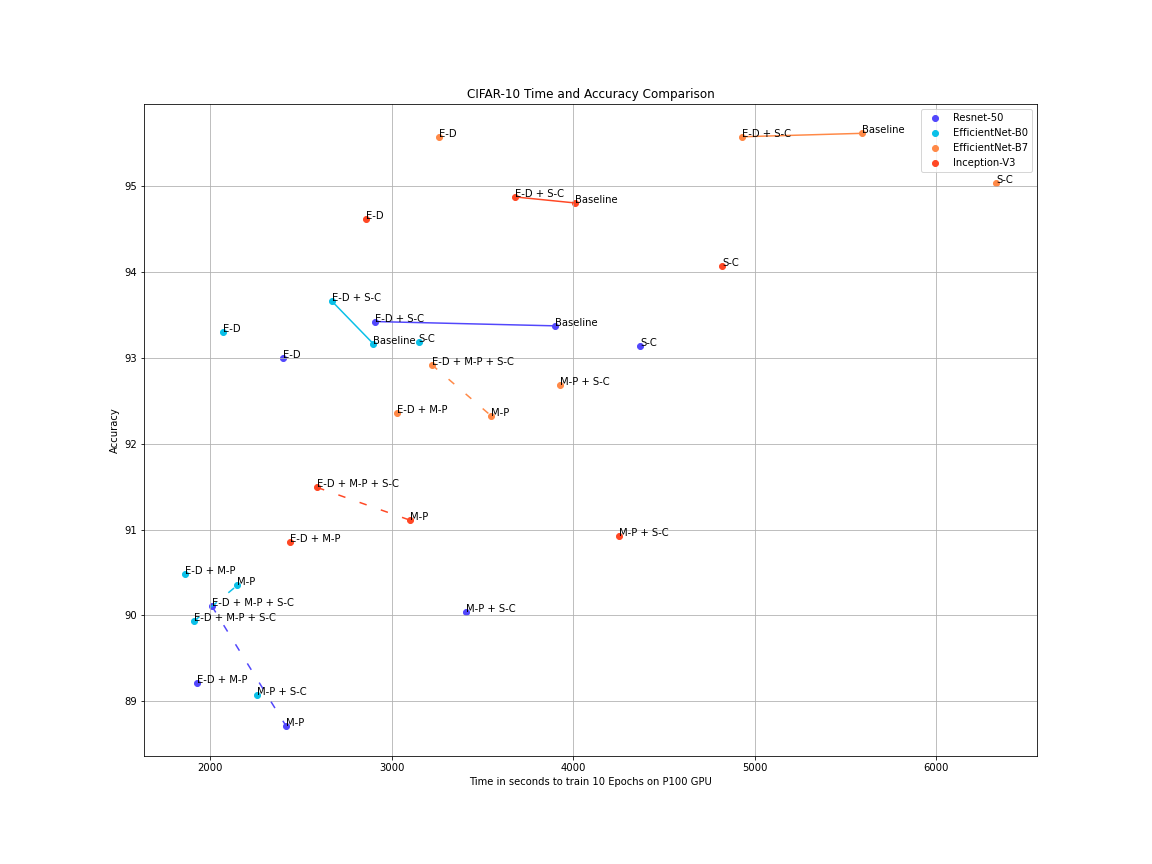}
  \caption{Comparison of Time and Accuracy for 10 Epochs on CIFAR-10. E-D represents Encoding-Decoding Optimization pipeline, M-P represents Mixed-Precision Optimization pipeline, whereas, S-C represents Sequential-Checkpoints Optimization pipeline. X-axis represents Time taken for 10 Epochs and Y-axis represents Accuracy. Line between Baseline and E-D + S-C represents difference in achieved accuracy and time taken to train for 10 Epochs. Dashed Line between M-P and E-D + M-P + S-C represents difference in achieved accuracy and time taken to train for 10 Epochs using Mixed Precision.} 
  \label{fig:acc_time}
\end{figure*}

\begin{figure*}
  \includegraphics[width=\textwidth,]{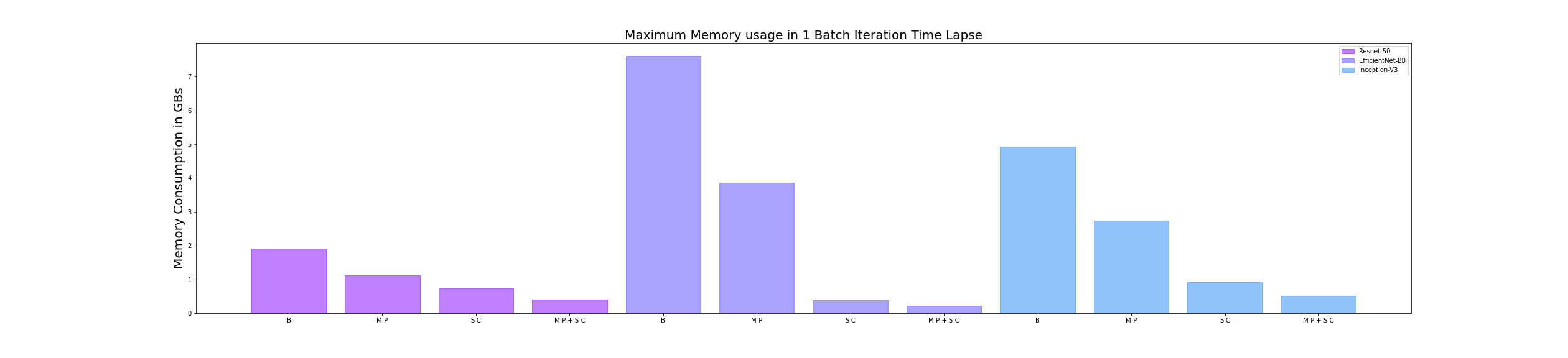}
  \caption{Memory Consumption Comparison of famous Image models and multiple optimization pipelines for 1 Batch iteration. B represents standard baseline pipeline, E-D represents Encoding-Decoding Optimization pipeline, M-P represents Mixed-Precision Optimization pipeline, whereas, S-C represents Sequential-Checkpoints Optimization pipeline. X-axis represents optimization pipelines and Y-axis represents memory consumption in GBs.} 
  \label{fig:mem_cons}
\end{figure*}

Figure \ref{fig:acc_time} represents detailed comparison of widely used image models with multiple optimization techniques w.r.t accuracy and time taken for training. All these models are trained on P100 GPU with a batch size of 16. Resnet 50 trained for 10 epochs using standard pipeline achieved 93.3\% accuracy in approximately 3800 seconds whereas, sequential checkpoints achieved almost same accuracy in 4400 seconds. We can infer from our experiments that all networks trained with checkpoint optimization achieved almost same accuracy as standard pipeline but take more time to train. However, Figure \ref{fig:mem_res} and Figure \ref{fig:mem_cons} show that checkpoint optimization will consume much less memory than other pipelines. For example,  sequential checkpoints method reduced more than 50\% memory for Resnet 50 compared to standard baseline pipeline (Figure \ref{fig:mem_cons}).
These image models (Resnet 50, EfficientNet-b0, Inception-v3) trained by combining parallel encoding-decoding (E-D) and sequential checkpoints (S-C) optimization achieved same results in terms of accuracy as a baseline in less time and much less memory consumption. This combination can be represented as most optimized FP32 pipeline w.r.t memory, time and accuracy. Mixed-Precision combined with E-D, and S-C optimization takes even less time to train and less memory. 

OpTorch provides features to combine these pipelines easily with just a single command.

\textbf{from optorch import sc}

\textbf{scmodel = sc(model)}

We analyze the memory consumption of famous image models by using multiple optimization pipelines. Each bar in Figure \ref{fig:mem_cons} represents memory consumption in GBs. We performed this experiment by passing 1 batch of size 16 images of size (512x512x3). Resnet 50 standard pipeline consumes 2 GB, mixed precision consumes 1 GB, sequential checkpoints consumes 0.8 GB, and sequential checkpoints with mixed precision consumes 0.4 GB.

Sequential-Checkpoint optimization pipeline combined with E-D pipeline allows us to train in much less memory in some cases 10 times less memory than standard pipelines as shown in Figure \ref{fig:mem_cons}.

\section{Discussion \& Recommendation}
Due to memory limitations, architectures of the neural networks are constrained. One possible solution is to utilize more high power GPUs but the other one is to optimize implementations. Depth of neural network is directly proportional to the extra memory consumption while training. Standard implementations will save output of each intermediate layer for back-propagation. Hence, more number of layers means more memory consumption. 

Checkpoint optimization pipeline solves the problem of extra memory consumption but trades-off with extra time to train. E-D data flow pipeline combined with Checkpoint optimization gradient flow pipeline solves the issue of trade-off between extra memory consumption and extra time to train a neural network.

\begin{figure}
  \includegraphics[width=0.5\textwidth,]{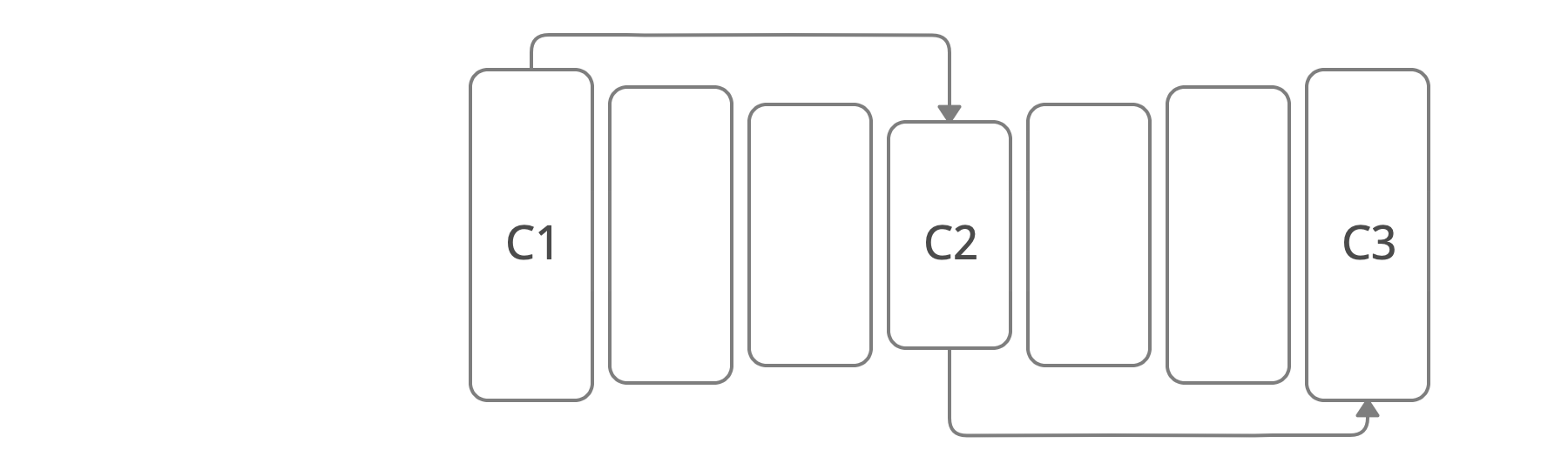}
  \caption{Best checkpoints in a simple neural network of 7 layers. C1 represents first checkpoint as we have to store input, C3 represents output of neural network that will be saved to evaluate loss while training. C2 represents checkpoint at intermediate layer.} 
  \label{fig:recom}
\end{figure}

Consider a simple neural network of 7 layers as shown in Figure \ref{fig:recom}. It shows the most optimized way to create checkpoints is to design a middle layer with less parameters. In this way checkpoint will be on small layer and it will consume less memory than other checkpoints.

We recommend designing neural network in Auto-encoder type or UNet type architectures as shown in Figure \ref{fig:recom} where we have small intermediate layer that can be used as a optimal checkpoint. It will result in much less memory consumption than other possible architectures.

\section{Conclusion}
In this paper, we propose OpTorch, a
machine learning library designed to overcome weaknesses in
existing implementations of neural network training. OpTorch
provides features to train complex neural networks with limited
computational resources. OpTorch achieved the same accuracy
as existing libraries on Cifar-10 and Cifar-100 datasets while
reducing memory usage to approximately 50\%. In our experiments, parallel encoding-decoding along
with sequential checkpoints result in much improved memory
and time usage while keeping the accuracy similar to existing
pipelines.

\section*{Acknowledgment}

%Dr. Reveryrand would like to acknowledge the funding by XLIM, Limoges, France. 
This work was supported by the National Center in Big Data and Cloud Computing (NCBC) and the National University of Computer and Emerging Sciences (NUCES-FAST), Islamabad, Pakistan.

% if have a single appendix:
%\appendix[Proof of the Zonklar Equations]
% or
%\appendix  % for no appendix heading
% do not use \section anymore after \appendix, only \section*
% is possibly needed

% use appendices with more than one appendix
% then use \section to start each appendix
% you must declare a \section before using any
% \subsection or using \label (\appendices by itself
% starts a section numbered zero.)
%

% ============================================
%\appendices
%\section{Proof of the First Zonklar Equation}
%Appendix one text goes here %\cite{Roberg2010}.

% you can choose not to have a title for an appendix
% if you want by leaving the argument blank
%\section{}
%Appendix two text goes here.

% use section* for acknowledgement
%\section*{Acknowledgment}

%The authors would like to thank D. Root for the loan of the SWAP. The SWAP that can ONLY be usefull in Boulder...

% Can use something like this to put references on a page
% by themselves when using endfloat and the captionsoff option.
\ifCLASSOPTIONcaptionsoff
  \newpage
\fi

% trigger a \newpage just before the given reference
% number - used to balance the columns on the last page
% adjust value as needed - may need to be readjusted if
% the document is modified later
%\IEEEtriggeratref{8}
% The "triggered" command can be changed if desired:
%\IEEEtriggercmd{\enlargethispage{-5in}}

% ====== REFERENCE SECTION

%\begin{thebibliography}{1}

% IEEEabrv,

\bibliographystyle{IEEEtran}
\bibliography{IEEEabrv}

\vfill

% Can be used to pull up biographies so that the bottom of the last one
% is flush with the other column.
%\enlargethispage{-5in}

% that's all folks
\end{document}